\documentclass[letterpaper, 10 pt, conference]{ieeeconf}  

\usepackage{graphicx}
\usepackage{amsmath}
\usepackage{bm}
\usepackage{cite}
\usepackage{flushend}
\include{preamble}

\IEEEoverridecommandlockouts                              

\title{\LARGE \textbf
  {
      LunaDrive: A Delay-Compensated High-Voltage GaN FET-Based Motor Driver for Dynamic Robots with Flat BLDC Motors
  }
}

\author{Sota Yuzaki$^{1}$, Temma Suzuki$^{1}$, Hiromi Tada$^{1}$, Masanori Konishi$^{1}$, Kento Kawaharazuka$^{1,2}$, and Kei Okada$^{1}$
  \thanks{$^{1}$ The authors are with the Department of Mechano-Informatics, Graduate School of Information Science and Technology, The University of Tokyo, 7-3-1 Hongo, Bunkyo-ku, Tokyo, 113-8656, Japan.
    {\texttt\small [yuzaki, t-suzuki, h-tada, konishi, kawaharazuka, k-okada]@jsk.t.u-tokyo.ac.jp}
  }
  \thanks{$^{2}$ The author is with the AI Center, Graduate School of Information Science and Technology, The University of Tokyo, Japan.}
}
\begin{document}

\maketitle
\thispagestyle{empty}
\pagestyle{empty}

\begin{abstract}

The performance improvement of high-power flat BLDC motors has accelerated the development of dynamic robots. However, many commercially available servo motors assume operating voltages of \SI{48}{\volt} or lower, which limits the maximum rotational speed. Dynamic robots require rapid energy generation, so this voltage constraint restricts motion performance. Therefore, driving motors beyond the rated voltage is desirable to increase the instantaneous maximum speed.
On the other hand, semiconductor devices used in motor drivers have a trade-off between voltage rating and current capacity. Conventional drivers using Si MOSFETs have difficulty achieving both high-voltage and high-current operation. Although GaN FETs are promising, compact drivers that can be mounted on the rear side of flat BLDC motors remain limited.
In this study, a motor driver for high-power flat BLDC motors using GaN FETs is developed. The effect of delay compensation in the high-speed region beyond the rated operating range is also investigated.
In the experiments, under \SI{96}{\volt} operation, a continuous current of \SI{30}{A} was achieved with a heat sink attached. A peak current of \SI{80}{A} and a maximum electrical frequency of \SI{3110}{\hertz} were confirmed. A high-speed load lifting experiment driven by a LiPo battery 24S (\SI{100}{\volt}) was also conducted, demonstrating applicability to dynamic robot operation.
\end{abstract}

\section{Introduction}\label{sec:introduction}
In recent years, robot motion performance has significantly improved, and dynamic motions such as backflips by humanoid and quadruped robots have been widely demonstrated \cite{unitree_h1_backflip, mini_cheetah}. These robots commonly use high-power flat BLDC motors, which provide high torque and enable low gear ratio mechanisms suitable for dynamic motion. Recently, integrated servo motors combining a motor, gearbox, and motor driver have also become widely available and are adopted in many robots because of their convenience \cite{mevius, duke_humanoid, backdrivable_arm}.

However, many of these servo motors operate at \SI{48}{\volt} or lower \cite{robstride, ak}. Since motor speed is proportional to the applied voltage, this voltage effectively limits the maximum rotational speed and causes speed saturation during high-speed motion. Therefore, dynamic robots require operation beyond the rated motor voltage.

Semiconductor devices used in motor drivers, however, exhibit a trade-off between voltage rating and current capability, making sufficient current difficult to supply at high voltage. Because semiconductor performance limits depend on material properties \cite{semiconductor_figure}, changing the material is an effective approach. Replacing the Si MOSFETs used in many motor drivers with high-performance GaN FETs enables both high-voltage and high-current operation. Accordingly, GaN FET-based BLDC motor drives have been actively studied in power electronics \cite{gan_md_modeling_experiment, gan_md_efficiency_size, gan_md_high_power}.

Although these studies demonstrate the effectiveness of GaN FETs, many target experimental benches or industrial applications and are difficult to integrate directly into robots because of size and weight constraints. Few drivers offer the compactness and high power density required for mounting behind high-power flat BLDC motors. We therefore designed LunaDrive, a compact GaN FET motor driver for this configuration, under severe electrical, thermal, and packaging constraints.

\begin{figure}[t]
  \centering
  \includegraphics[width=1.0\columnwidth]{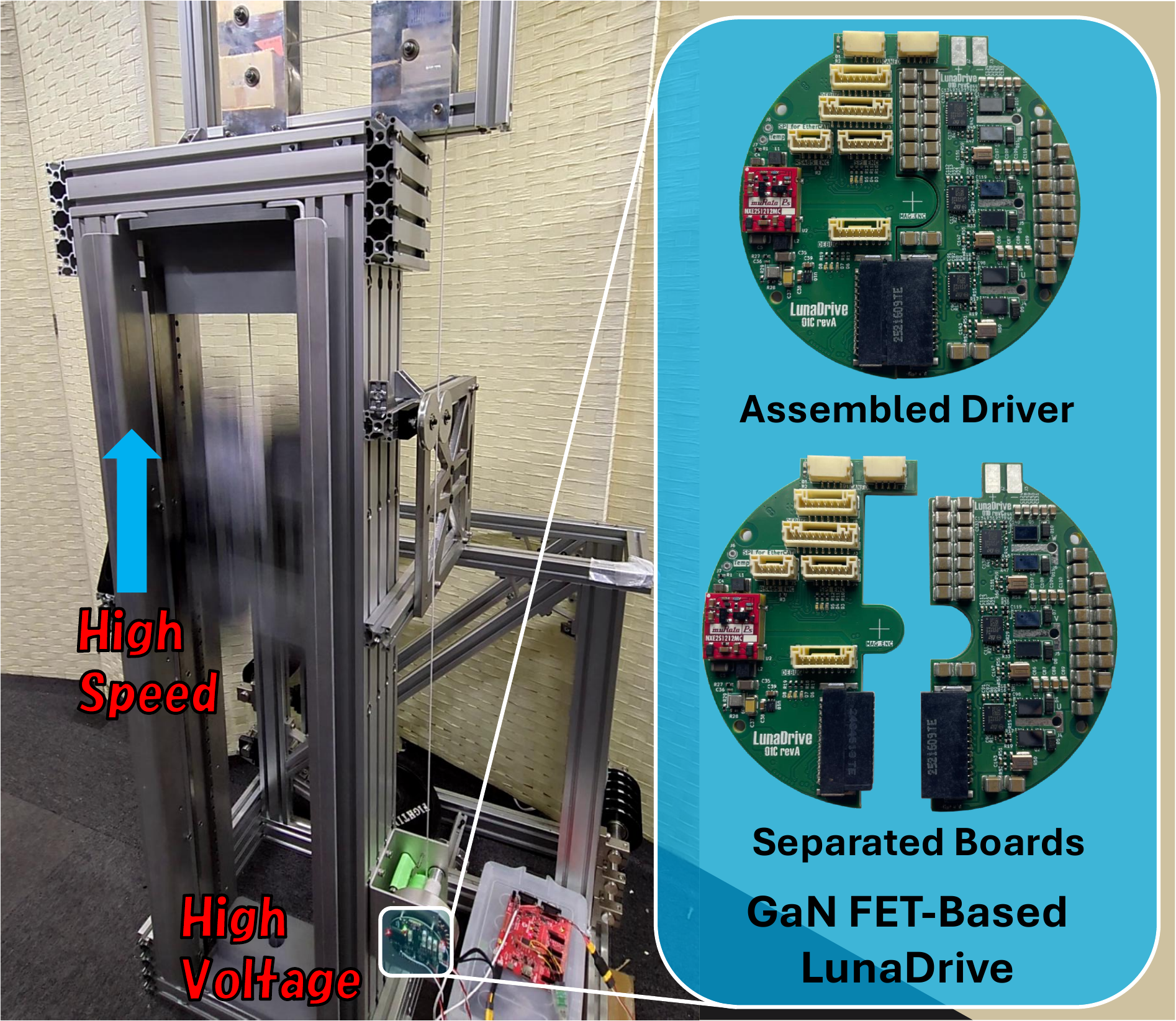}
  \vspace{-1.5em}
  \caption{LunaDrive enabling high-voltage, high-speed operation for dynamic robots}
  \label{figure:fig1}
  \vspace{-1.0em}
\end{figure}

The previous discussion focused on hardware, but software is also important. Control of a BLDC motor requires the electrical angle, which is given by $p$ times the mechanical angle. Here, $p$ denotes the number of pole pairs. The high-power flat BLDC motors considered in this study typically have a large pole pair number of $p=21$. When driven beyond the rated voltage, the electrical angular velocity becomes very high. Under the conditions assumed in this study, the maximum electrical frequency reaches approximately \SI{3000}{\hertz}. In contrast, the recommended maximum electrical frequency of commonly used BLDC motor drivers \cite{odrive_pro} is about \SI{700}{\hertz}. Therefore, this study targets a frequency range several times higher.

At such high electrical angular velocities, system delay increases the electrical-angle phase error and may degrade or destabilize control. Although delay compensation in motor control systems has been widely studied \cite{delay_compensation_deat_beat, delay_compensation_offset_and_time, delay_compensation_fourier}, most studies consider relatively low electrical frequencies. We experimentally evaluate delay compensation at high electrical frequencies using the developed driver.

\begin{figure}[t]
  \centering
  \includegraphics[width=0.9\columnwidth]{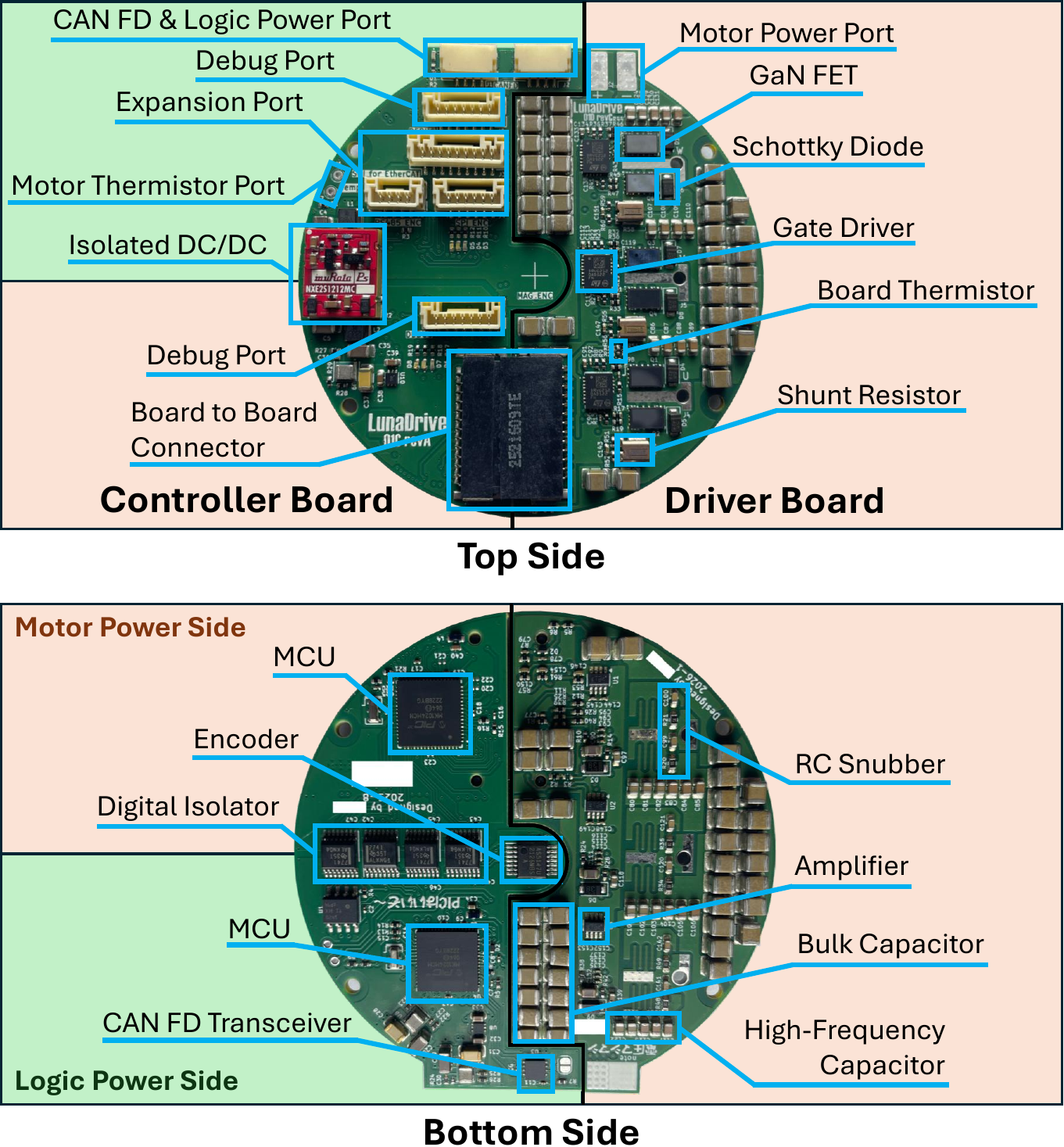}
  \vspace{-1.0em}
  \caption{LunaDrive composed of two boards: a controller board and a driver board}
  \label{fig:circuit_picture}
\end{figure}

\begin{figure}[t]
  \centering
  \includegraphics[width=1.0\columnwidth]{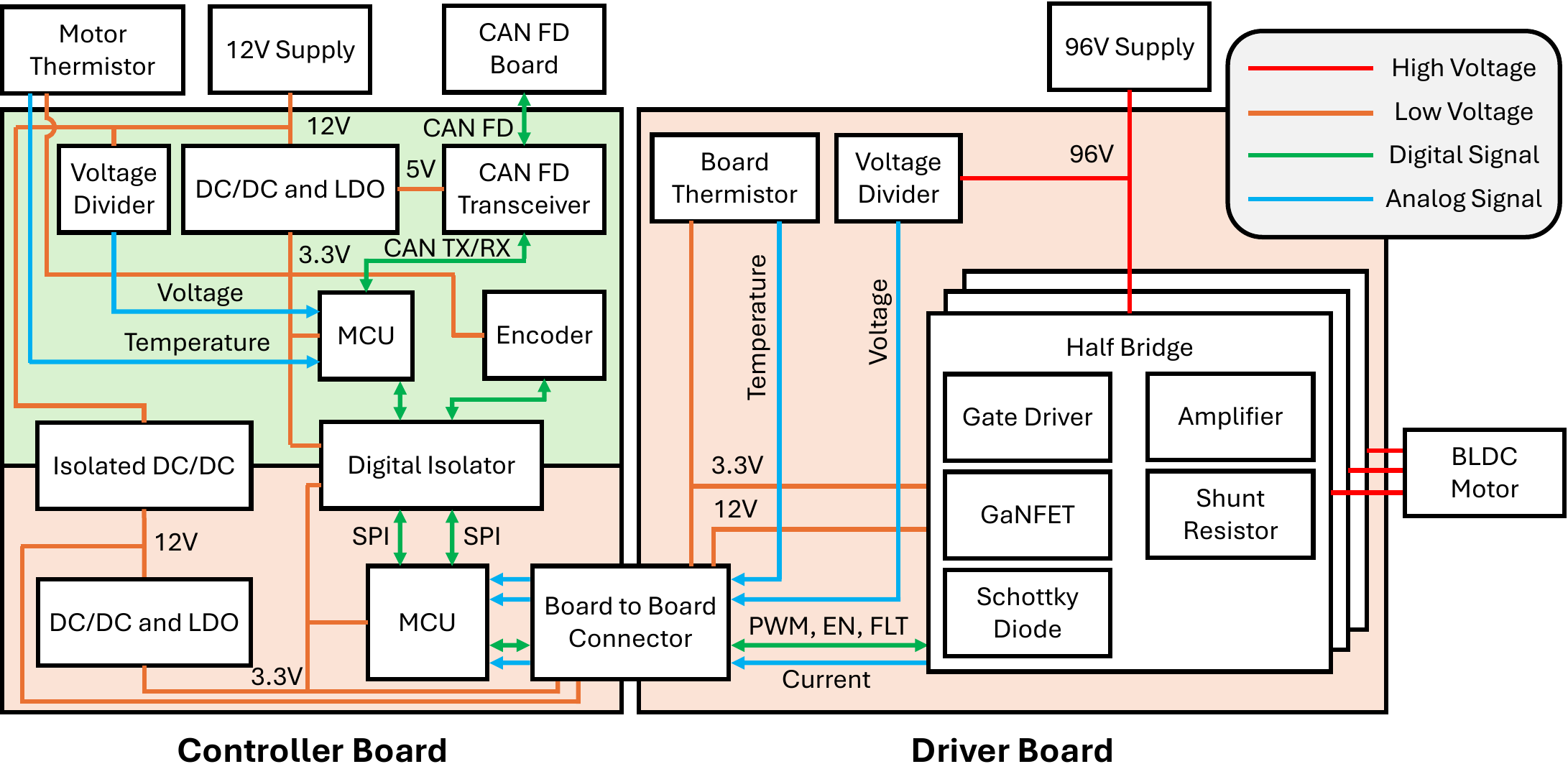}
  \vspace{-1.5em}
  \caption{Electronic block diagram of LunaDrive}
  \label{fig:circuit_block_diagram}
  \vspace{-1.5em}
\end{figure}

\begin{figure}[t]
  \centering
  \includegraphics[width=1.0\columnwidth]{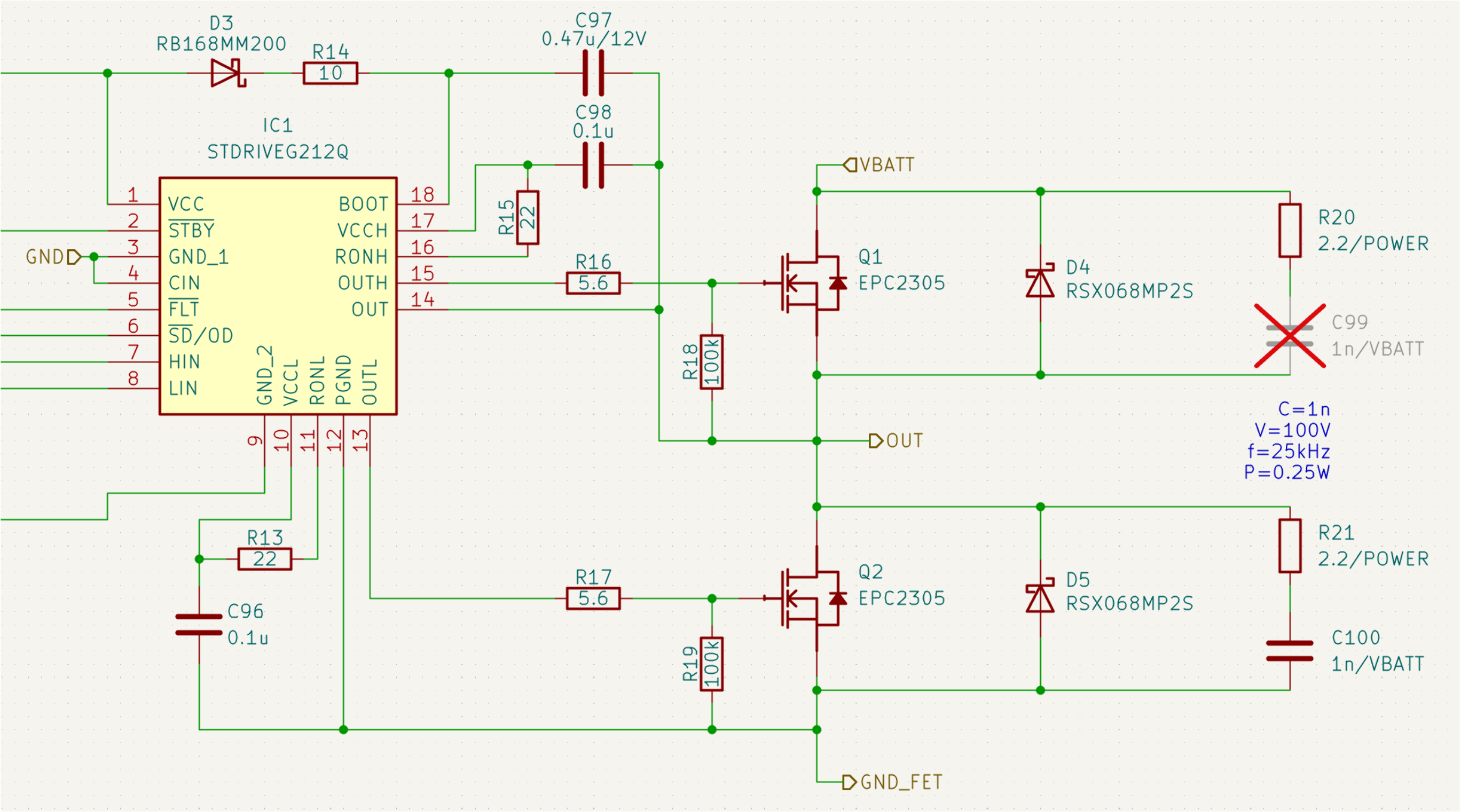}
  \vspace{-1.0em}
  \caption{Schematic of GaN FET-based half-bridge circuit}
  \label{fig:halfbridge_schematic}
\end{figure}

\begin{table}[t]
    \centering
    \caption{Performance comparison of 150-V power FETs}
    \label{tab:fet}
    \begin{tabular}{cccc}
        \hline
        Device & Type & On-Resistance & Size \\
        & & [\si{\milli\ohm}] & [\si{mm^2}] \\
        \hline
        FET-1 & GaN FET & 3 & 15 \\
        FET-2 & MOSFET & 2.5 & 116 \\
        FET-3 & MOSFET & 5.6 & 30 \\
      \hline
    \end{tabular}
    \par\medskip
    \footnotesize
    FET-1: EPC2305 (EPC).  
    FET-2: IPT025N15NM6ATMA1 (Infineon Technologies).
    FET-3: SIRS5700DP-T1-RE3 (Vishay Siliconix).
    \vspace{-1.5em}
\end{table}

As an application example of LunaDrive, a wire-driven robot that winds a wire using a motor as a power transmission mechanism is considered. Robots that achieve large jumps by securing a long acceleration distance with a long wire have been reported \cite{ramiel}. In such systems, speed saturation becomes a design bottleneck, and operation beyond the rated voltage is required. Therefore, a wire module using LunaDrive is developed and its effectiveness for dynamic robots is evaluated (\figref{figure:fig1}).

\begin{table*}[tb]
    \centering
    \caption{Selected high-areal-capacitance MLCCs across package form factors and voltage ratings\\evaluated under practical DC bias conditions}
    \label{tab:capacitor}
    \begin{tabular}{cccccccc}
        \hline
        Manufacturer & Part Number &
        Nominal Cap. & Retention (DC Bias) &
        Rated Volt. & Package Size &
        Height & Areal Cap. \\
        & &
        [$\si{\micro\farad}$] &
        [\%] &
        [$\si{\volt}$] &
        &
        [$\si{\milli\meter}$] &
        [$\si{\micro\farad\per\milli\meter\squared}$] \\
        \hline
        Murata & GRM32EC72A106KE05 & 10 & 25.1 & 100 & 3225M & 2.5 & 0.078 \\
        TDK & C5750X7S2A226M280KB & 22 & 45.5 & 100 & 5750M & 3.1 & 0.088 \\
        TDK & CAA572X7S2A336M640LJ & 33 & 47.0 & 100 & 6050M & 6.9 & 0.129 \\
        TDK & C3225X7T2E334K200AA & 0.33 & 69.3 & 250 & 3225M & 2.0 & 0.029 \\
        TDK & C5750X7T2E225K250KA & 2.2 & 58.5 & 250 & 5750M & 2.5 & 0.045 \\
        TDK & CKG57NX7T2E335M500JJ & 3.3 & 60.5 & 250 & 6050M & 5.5 & 0.067 \\
      \hline
    \end{tabular}
\end{table*}

\begin{table*}[tb]
    \centering
    \caption{Capacitors mounted on the power supply line of LunaDrive}
    \label{tab:capacitor_mounted}
    \begin{tabular}{ccccccc}
        \hline
        Manufacturer & Part Number &
        Nominal Cap. & Rated Volt. &
        Package Size & Height & Quantity \\
        & &
        [$\si{\micro\farad}$] &
        [$\si{\volt}$] &
        &
        [$\si{\milli\meter}$] &
        [-] \\
        \hline
        Murata & GRM32EC72A106KE05 & 10 & 100 & 3225M & 2.5 & 76 \\
        TDK & C2012X7T2E104K125AA & 0.1 & 250 & 2012M & 1.25 & 18 \\
        TDK & CGA4J3X7R2E223K125AE & 0.022 & 250 & 2012M & 1.25 & 12 \\
      \hline
    \end{tabular}
    \vspace{-1em}
\end{table*}

\section{Method} \label{sec:method}
\subsection{Design of LunaDrive with GaNFET-Based Driving}
LunaDrive consists of two boards, as shown in \figref{fig:circuit_picture}. One is a controller board for motor control and communication with the host system. The other is a driver board that drives the BLDC motor using GaN FETs. The diameter is \SI{70}{mm}. Only the connector section used by the user protrudes outward. The thickness is \SI{8.51}{mm}. The power system has two supplies: a logic supply (\SI{12}{V}) for the microcontroller and gate drive circuits, and a motor supply (\SI{96}{V}, peak \SI{150}{V}). The power stage components are rated for \SI{150}{V}.

\subsubsection{Design of the controller board}
The configuration of the controller board is shown on the left side of \figref{fig:circuit_block_diagram}. The board is separated into logic and motor power domains by an isolated DC/DC converter and digital isolators. A 6-layer PCB with 1 oz copper was used to improve implementation density and noise immunity.

The logic power domain includes DC/DC converters, a Low Dropout Regulator (LDO), a microcontroller for communication and higher-level control, a magnetic encoder for motor angle measurement, a CAN FD transceiver, a voltage divider for supply voltage measurement, and a thermistor for motor temperature measurement. PIC32MK1024MCM064 (Microchip), which integrates CAN FD and provides abundant PWM and ADC peripherals, was used as the microcontroller. The motor power domain similarly consists of a power circuit and a microcontroller for motor control.

\subsubsection{Design of the driver board}
The configuration of the driver board is shown on the right side of \figref{fig:circuit_block_diagram}. The driver board consists of a voltage divider for supply voltage measurement, a thermistor for board temperature measurement, and three GaN FET-based half-bridge circuits for motor driving. To handle large currents, a 2 oz 8-layer PCB (total 16 oz copper) was used.

The GaN FET-based half-bridge is shown in \figref{fig:halfbridge_schematic}. As summarized in \tabref{tab:fet}, GaN FETs are smaller than Si MOSFETs with similar on-resistance and have lower on-resistance than similarly sized Si MOSFETs. We selected the EPC2305 (EPC), which has a breakdown voltage of \SI{150}{\volt}, an on-resistance of \SI{3}{\milli\ohm}, and dimensions of \SI{3}{\milli\meter}$\times$\SI{5}{\milli\meter}.

\textbf{Dead time.} Compared with Si MOSFETs, GaN FETs have higher reverse voltage and tend to increase reverse conduction loss during dead time, although they allow high-speed switching. Previous studies reduce this loss by shortening the dead time \cite{gan_reverse_conduction, gan_dead_time}. In this study, stable operation under motor load with large current fluctuations was prioritized. Therefore, a relatively long dead time of \SI{100}{ns} was adopted, and a Schottky barrier diode was connected in parallel to suppress reverse conduction loss.

\textbf{Gate drive.} In Si MOSFETs, the gate voltage rating is typically \SI{20}{V} and the recommended gate voltage is about \SI{12}{V}. In contrast, GaN FETs typically have a gate voltage rating of \SI{6}{V} and a recommended gate voltage of \SI{5}{V}, leaving only a \SI{1}{V} margin and making gate driving difficult. In this study, STDRIVEG212 (STMicroelectronics), a gate driver with integrated LDOs for both low-side and high-side circuits, was adopted. The wiring length from the driver to the GaN FET was designed to be shorter than \SI{5}{mm}.

\textbf{Capacitor selection method.} To handle large currents on a compact PCB, sufficient effective capacitance must be secured within a limited mounting area. When high-voltage Multilayer Ceramic Capacitors (MLCCs) are connected in series, the effective capacitance decreases in proportion to $1/N_s$, while the mounting area increases in proportion to $N_s$. As a result, the capacitance efficiency per unit area decreases in proportion to $1/N_s^2$.

To evaluate this efficiency, the effective capacitance per unit area $C_\mathrm{area}$ with DC bias reduction is defined as follows.

\begin{equation}
C_\mathrm{area}
=
\frac{
C_\mathrm{nom}\, r_\mathrm{DC}
}{
A\, N_s^{2}
}
\end{equation}

Here, $C_\mathrm{nom}$ is the nominal capacitance, $r_\mathrm{DC}$ is the capacitance retention ratio under DC bias, $A$ is the mounting area of a single device, and $N_s$ is the number of series connections.

\textbf{Capacitor evaluation.} The results are shown in \tabref{tab:capacitor}. Assuming a supply voltage of \SI{100}{\volt}, \SI{100}{\volt}-rated capacitors were evaluated with two devices in series, while \SI{250}{\volt}-rated capacitors were evaluated with one device. The DC bias values were taken at \SI{50}{\volt} and \SI{100}{\volt}, respectively. As a result, two \SI{100}{\volt}-rated capacitors in series provided higher effective capacitance per unit area than one \SI{250}{\volt}-rated capacitor. Within the same voltage rating, larger packages also showed higher capacitance per unit area.

\textbf{Capacitor selection for flat BLDC motors.} To keep the board thin, the capacitor height was restricted to \SI{4}{\milli\meter}. In addition, as shown in \figref{fig:circuit_picture}, the available mounting area is a semicircle excluding a rectangular region, so large packages may reduce placement efficiency. Considering these constraints, GRM32EC72A106KE05 (Murata, \SI{10}{\micro\farad}, \SI{100}{\volt}, 3225M) was adopted as shown in \tabref{tab:capacitor_mounted}. Additional capacitors were placed to suppress high-frequency noise from GaN FET switching. In total, 106 capacitors were mounted on the power line for high-current operation.

\begin{figure*}[t]
  \centering
  \includegraphics[width=2.0\columnwidth]{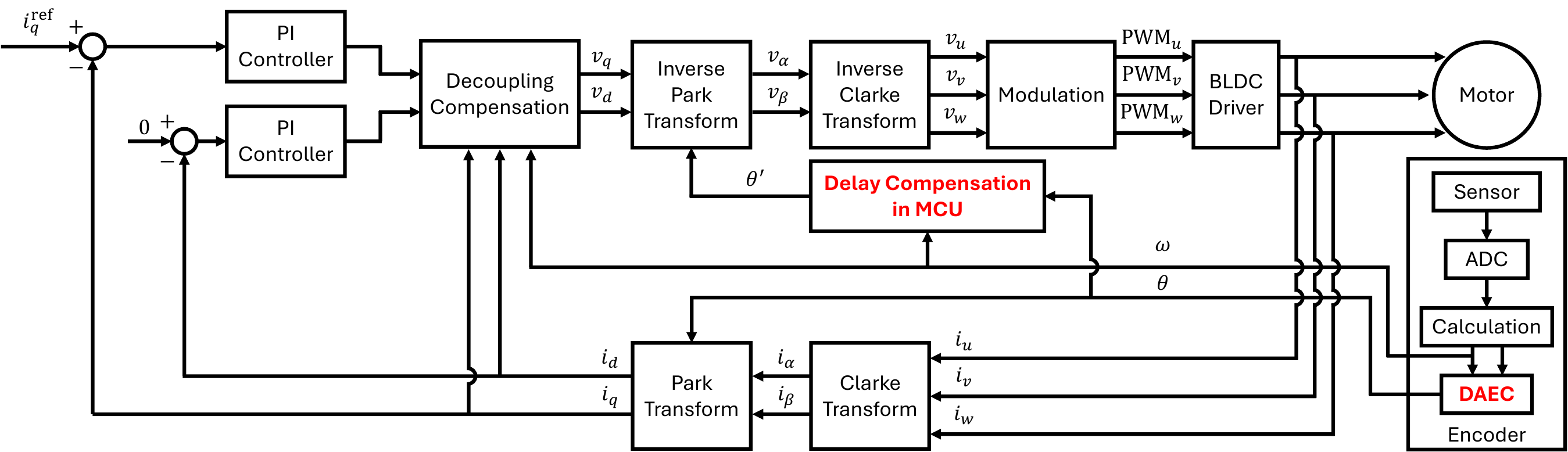}
  \vspace{-1em}
  \caption{Control block diagram of LunaDrive with delay compensation}
  \label{fig:control_block_diagram}
  \vspace{-1.5em}
\end{figure*} 

\begin{figure}[t]
  \centering
  \includegraphics[width=1.0\columnwidth]{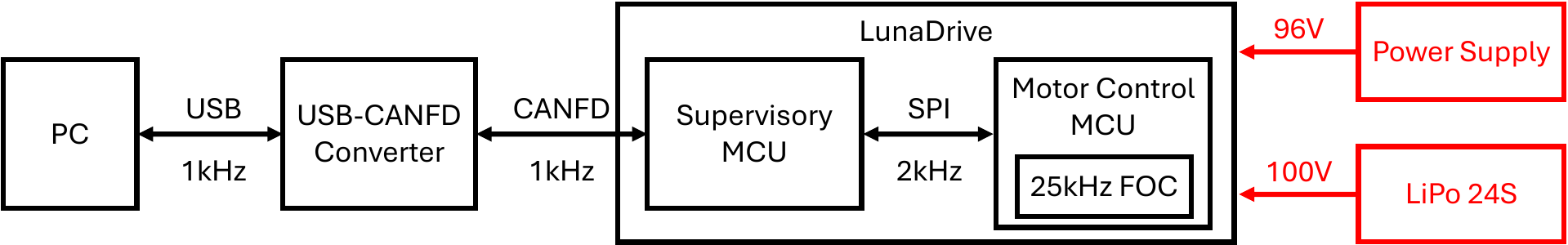}
  \vspace{-2em}
  \caption{Overall system configuration}
  \label{fig:system}
  \vspace{-1.5em}
\end{figure}

\subsection{Control of LunaDrive with Delay Compensation}
Field-Oriented Control (FOC) is used for BLDC motor control. In FOC, the three-phase AC quantities in the uvw coordinate system are transformed into the dq coordinate system by Clarke and Park transformations. Control is then performed on these quantities as DC values. An amplitude-invariant transformation was used so that the amplitudes of voltage and current are preserved.

The coordinate transformations in FOC and the voltage equations in the dq coordinate system are shown below.

\begin{equation}
  \begin{bmatrix}
    f_\alpha \\
    f_\beta \\
  \end{bmatrix}
  = \frac{2}{3}
  \begin{bmatrix}
    1 & -\frac{1}{2} & -\frac{1}{2} \\
    0 & \frac{\sqrt{3}}{2} & -\frac{\sqrt{3}}{2} \\
  \end{bmatrix}
    \begin{bmatrix}
    f_u \\
    f_v \\
    f_w \\
  \end{bmatrix}
\end{equation}

\begin{equation}
  \begin{bmatrix}
    f_d \\
    f_q \\
  \end{bmatrix}
  =
  \begin{bmatrix}
    \cos{\theta_e} & \sin{\theta_e} \\
    -\sin{\theta_e} & \cos{\theta_e} \\
  \end{bmatrix}
    \begin{bmatrix}
    f_\alpha \\
    f_\beta \\
  \end{bmatrix}
\end{equation}

\begin{equation}
  \begin{bmatrix}
    V_d \\
    V_q \\
  \end{bmatrix}
  =
  \begin{bmatrix}
    R & -L\omega_e \\
    L \omega_e & R \\
  \end{bmatrix}
  \begin{bmatrix}
    i_d \\
    i_q \\
  \end{bmatrix}
  + L \frac{d}{dt}
  \begin{bmatrix}
    i_d \\
    i_q \\
  \end{bmatrix}
  +
  \begin{bmatrix}
    0 \\
    K_e \omega_e \\
  \end{bmatrix}
\end{equation}

Here, $\theta_e$ denotes the electrical angle, $\omega_e$ denotes the electrical angular velocity, $R$ denotes the phase resistance, $L$ denotes the phase inductance, and $K_e$ denotes the back-EMF constant with respect to the electrical angular velocity.

Since the Park transformation includes rotation by the electrical angle $\theta_e$, an error between the electrical angle obtained from the sensor and the actual electrical angle degrades motor efficiency. If the error becomes large, sufficient torque cannot be generated. When the motor rotates at an electrical angular velocity $\omega_e$ and a system delay $T_d$ exists, the estimated electrical angle $\theta_e^{'}$ is expressed by the following equation. Therefore, a large angle deviation is expected in the region where the electrical angular velocity is high.

\begin{equation}
\theta_e^{'} = \theta_e + T_d \omega_e
\end{equation}

The system delay consists of encoder delay, microcontroller delay, and delay after the driver stage. The delay in the above equation can be calculated in the microcontroller and compensated by adding it to the electrical angle. However, the encoder delay is asynchronous with the microcontroller, so $T_d$ is not constant and depends on the encoder sampling timing. In this study, AS5147U (ams OSRAM) with Dynamic Angle Error Compensation (DAEC) was adopted to compensate the encoder delay internally. The delay after the microcontroller was compensated by the same calculation in the microcontroller.

The current control consists of PI control and a decoupling compensation term, as shown below.

\begin{equation}
  \begin{bmatrix}
    V_d \\
    V_q \\
  \end{bmatrix}
  =
  \begin{bmatrix}
    0 & -L\omega_e \\
    L \omega_e & 0 \\
  \end{bmatrix}
  \begin{bmatrix}
    i_d^{\mathrm{ref}} \\
    i_q^{\mathrm{ref}} \\
  \end{bmatrix}
  +
  \begin{bmatrix}
    0 \\
    K_e \omega_e \\
  \end{bmatrix}
  +
  \begin{bmatrix}
    u_d \\
    u_q \\
  \end{bmatrix}
\end{equation}

\begin{equation}
  \begin{bmatrix}
    u_d \\
    u_q \\
  \end{bmatrix}
  =
  K_p
  \begin{bmatrix}
    i_d^{\mathrm{ref}}-i_d \\
    i_q^{\mathrm{ref}}-i_q \\
  \end{bmatrix}
  +
  K_i \int
  \begin{bmatrix}
    i_d^{\mathrm{ref}}-i_d \\
    i_q^{\mathrm{ref}}-i_q \\
  \end{bmatrix}
  dt
\end{equation}

\begin{equation}
i_d^{\mathrm{ref}} = 0
\end{equation}

The obtained dq voltage commands are transformed back into three-phase voltages and output by PWM. In this study, Space Vector Modulation (SVM) was adopted to achieve a high modulation index.

For a surface-mounted permanent magnet motor, the motor torque $\tau$ is expressed using the torque constant $K_t$ with respect to the q-axis current as follows. However, in the region where the current is very large, this relationship does not strictly hold due to magnetic saturation.

\begin{equation}
\tau = K_t i_q
\end{equation}

The overall system configuration is shown in \figref{fig:system}. The PC and LunaDrive communicate at \SI{1}{\kilo\hertz} via a USB-CAN FD converter. The microcontrollers inside LunaDrive communicate via SPI at \SI{2}{\kilo\hertz}. The motor control microcontroller executes FOC at \SI{25}{\kilo\hertz}.

The power supply is normally provided by a power supply unit at \SI{96}{\volt}. However, in the high-speed load lifting experiment assuming a jump motion, regenerative operation occurs. Therefore, a LiPo battery was used. In this case, \SI{100}{\volt} is supplied immediately after charging.

\begin{figure}[t]
  \centering
  \includegraphics[width=0.6\columnwidth]{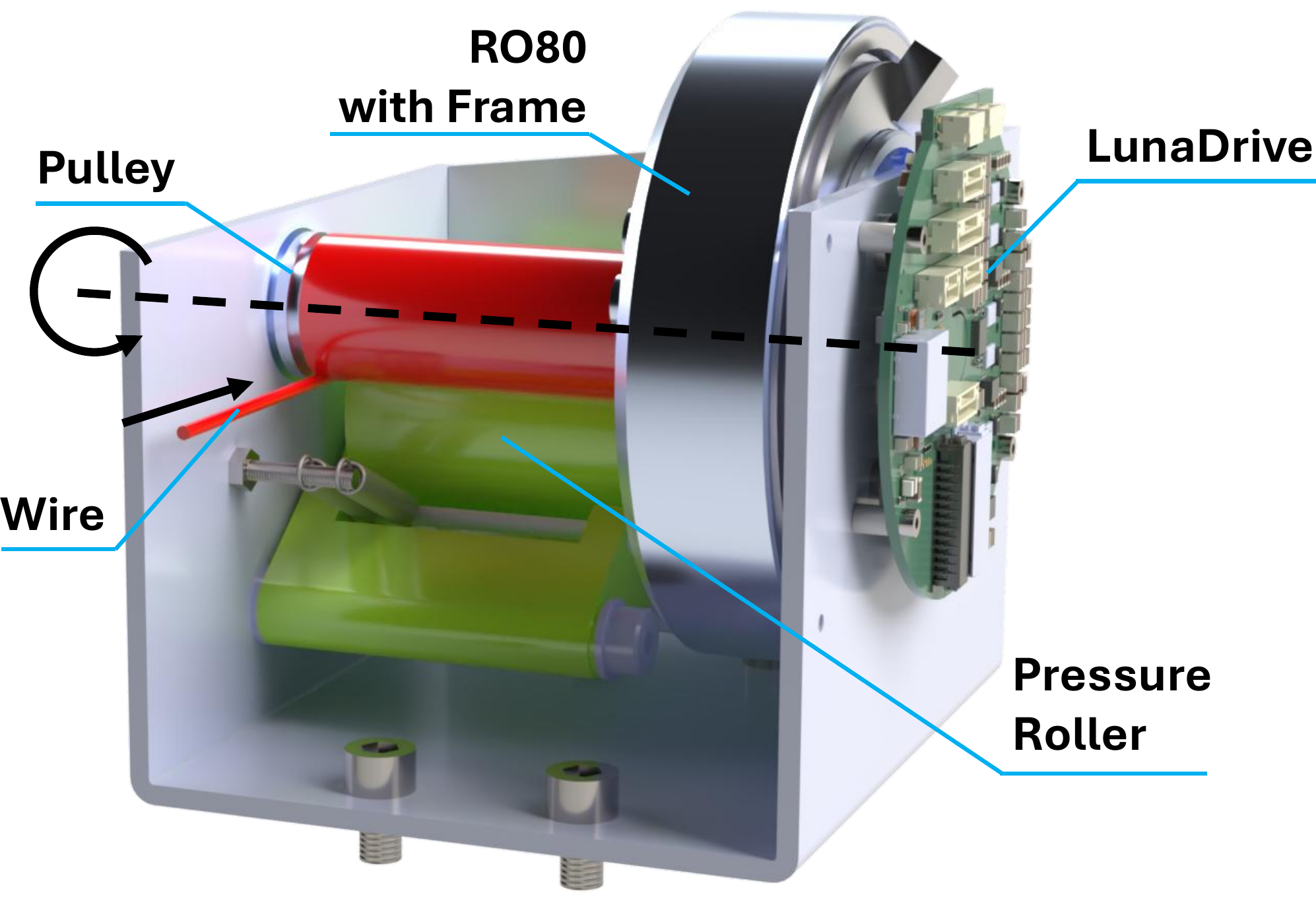}
  \caption{High-power wire module using LunaDrive}
  \label{fig:wire_module}
\end{figure}

\begin{table}[tb]
    \centering
    \caption{RO80 Frameless BLDC Motor Specifications}
    \label{tab:ro80}
    \begin{tabular}{cc}
        \hline
        Name & Value \\
        \hline
        Rated Voltage & \SI{48}{V} \\
        No Load Speed & \SI{5040}{rpm} \\
        Rated Torque & \SI{1.3}{Nm} \\
        Peak Torque & \SI{4}{Nm} \\
        Pole Pairs & 21 \\
      \hline
    \end{tabular}
    \vspace{-1.5em}
\end{table}

\subsection{Design of a High-Power Wire Winding Module Using LunaDrive}
Using the designed LunaDrive and the frameless BLDC motor RO80 (CubeMars) \cite{ro80_lite}, a high-power wire module shown in \figref{fig:wire_module} was developed. The main specifications of the RO80 are summarized in \tabref{tab:ro80}. The module consists of a pulley and a pressure roller.

The diameter of the pulley is \SI{18}{\milli\meter}. The diameter of the wire used, which has a Zylon core and a polyester sheath, is \SI{2}{\milli\meter}. Therefore, the winding radius is \SI{10}{\milli\meter}. Using this winding radius $r$, the wire tension $T$ is expressed by the following equation.

\begin{equation}
T = \frac{\tau}{r} 
\end{equation}

The pressure pulley presses the wire against the pulley to prevent slack when the wire is wound under no-load conditions.

\begin{figure}[t]
  \centering
  \includegraphics[width=0.8\columnwidth]{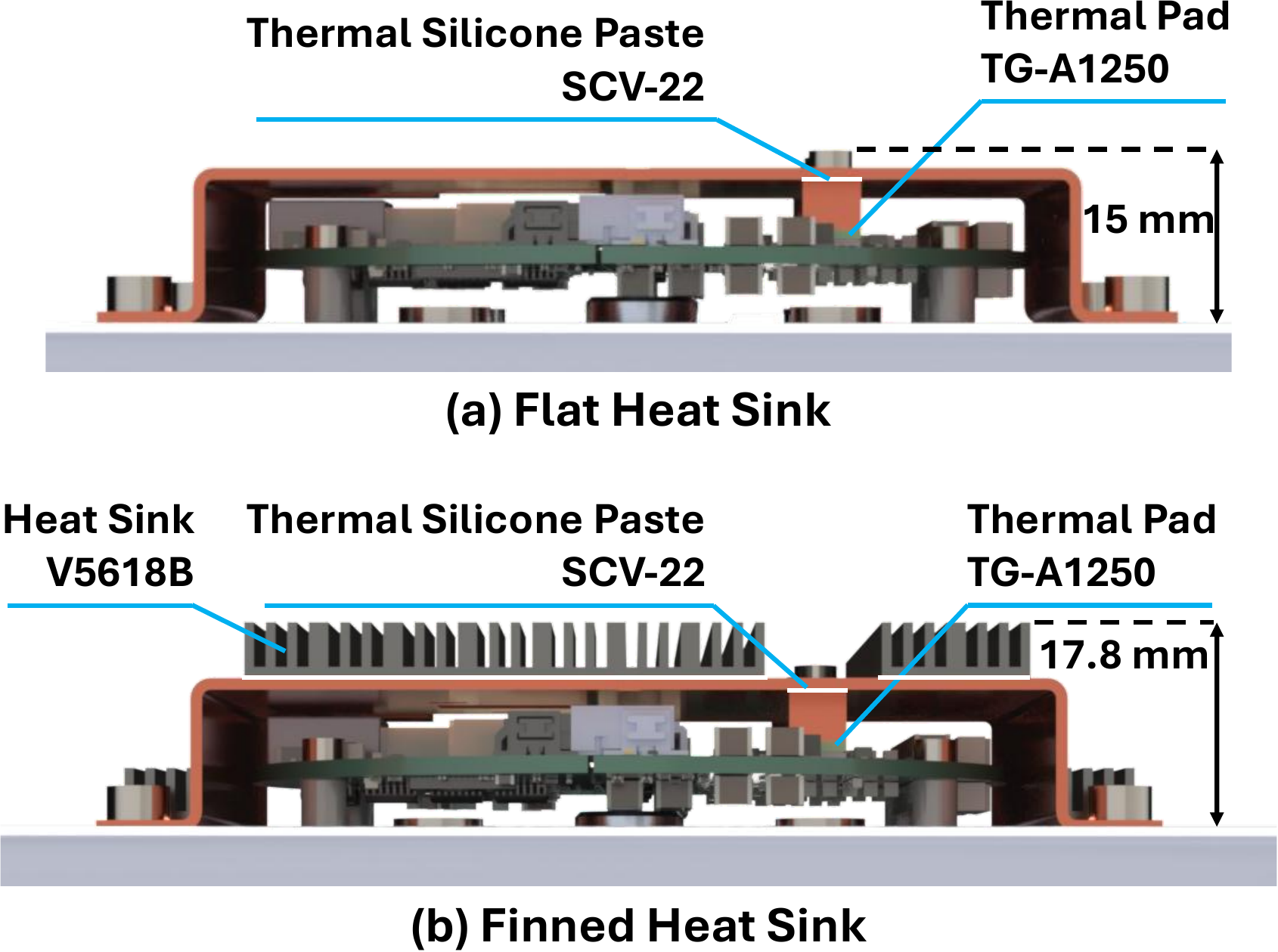}
  \caption{Two types of heat sinks}
  \label{fig:heat_sink}
\end{figure}

To reduce the thermal resistance of the GaN FETs used in the motor driver, two types of heat sinks were fabricated as shown in \figref{fig:heat_sink}. One is a Flat Heat Sink composed of Thermal Interface Material (TIM), a cover made of a \SI{1}{\milli\meter}-thick copper plate, and a machined copper component. The other is a Finned Heat Sink, in which 24 small commercial heat sinks are attached on the Flat Heat Sink using TIM. The heights from the frame are \SI{15}{\milli\meter} for the Flat Heat Sink and \SI{17.8}{\milli\meter} for the Finned Heat Sink.

\begin{figure}[t]
  \centering
  \includegraphics[width=0.6\columnwidth]{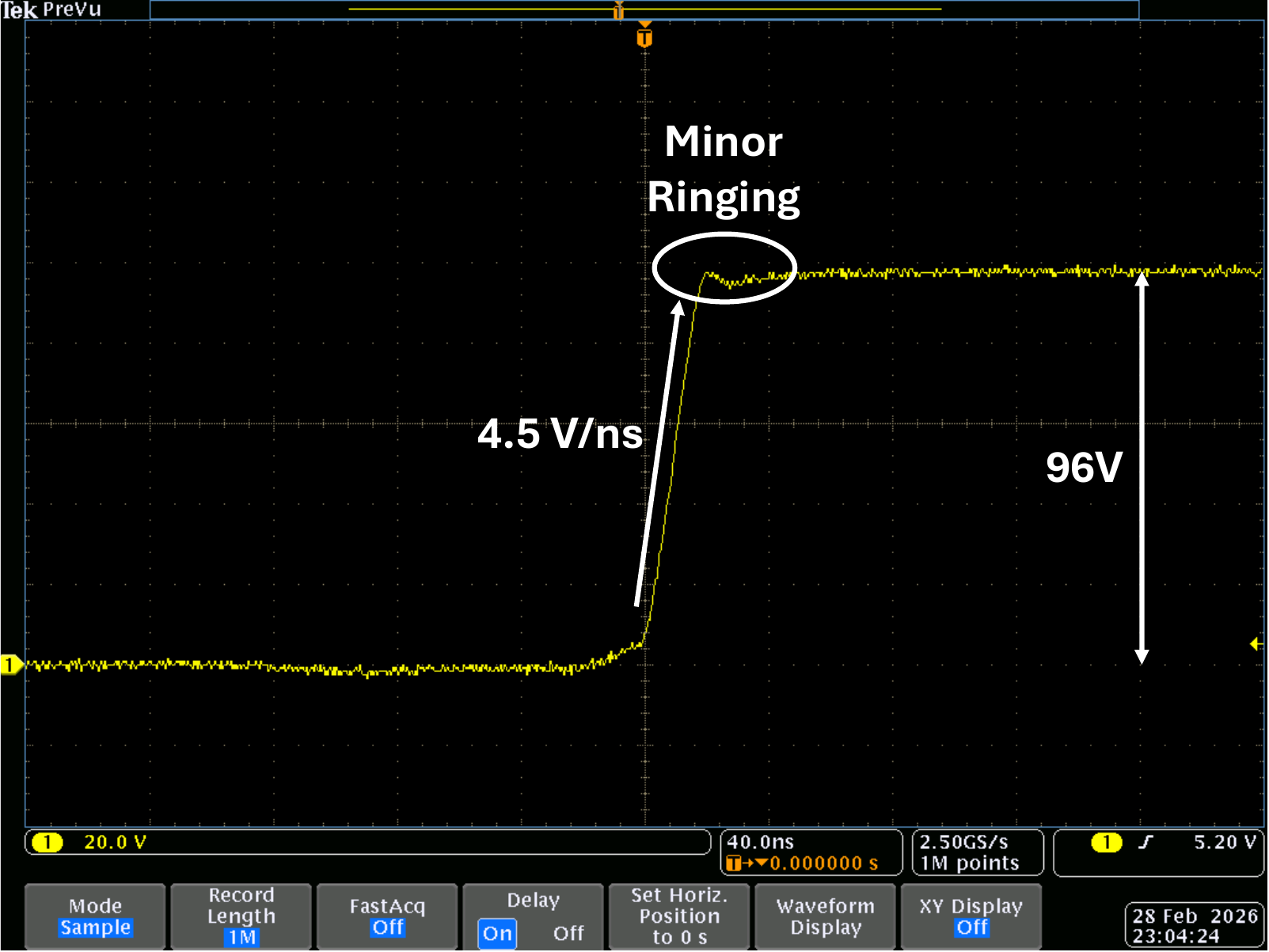}
  \caption{Switching waveform of drain-to-source voltage at 10 A motor current}
  \label{fig:waveform}
  \vspace{-1.5em}
\end{figure}

\begin{table}[t]
    \centering
    \caption{Results of the High-Speed Operation Experiment}
    \label{tab:speed}
    \begin{tabular}{cccc}
        \hline
        Voltage & Delay Compensation & Speed & Electrical Frequency \\
        {[V]} & & [rpm] & [Hz] \\
        \hline
        48 & Disabled & 4340 & 1520 \\
        48 & Enabled & 4340 & 1520 \\
        96 & Disabled & 5760 & 2020 \\
        96 & Enabled & 8890 & 3110 \\
      \hline
    \end{tabular}
    \vspace{-1.0em}
\end{table}

\begin{figure}[t]
  \centering
  \includegraphics[width=0.9\columnwidth]{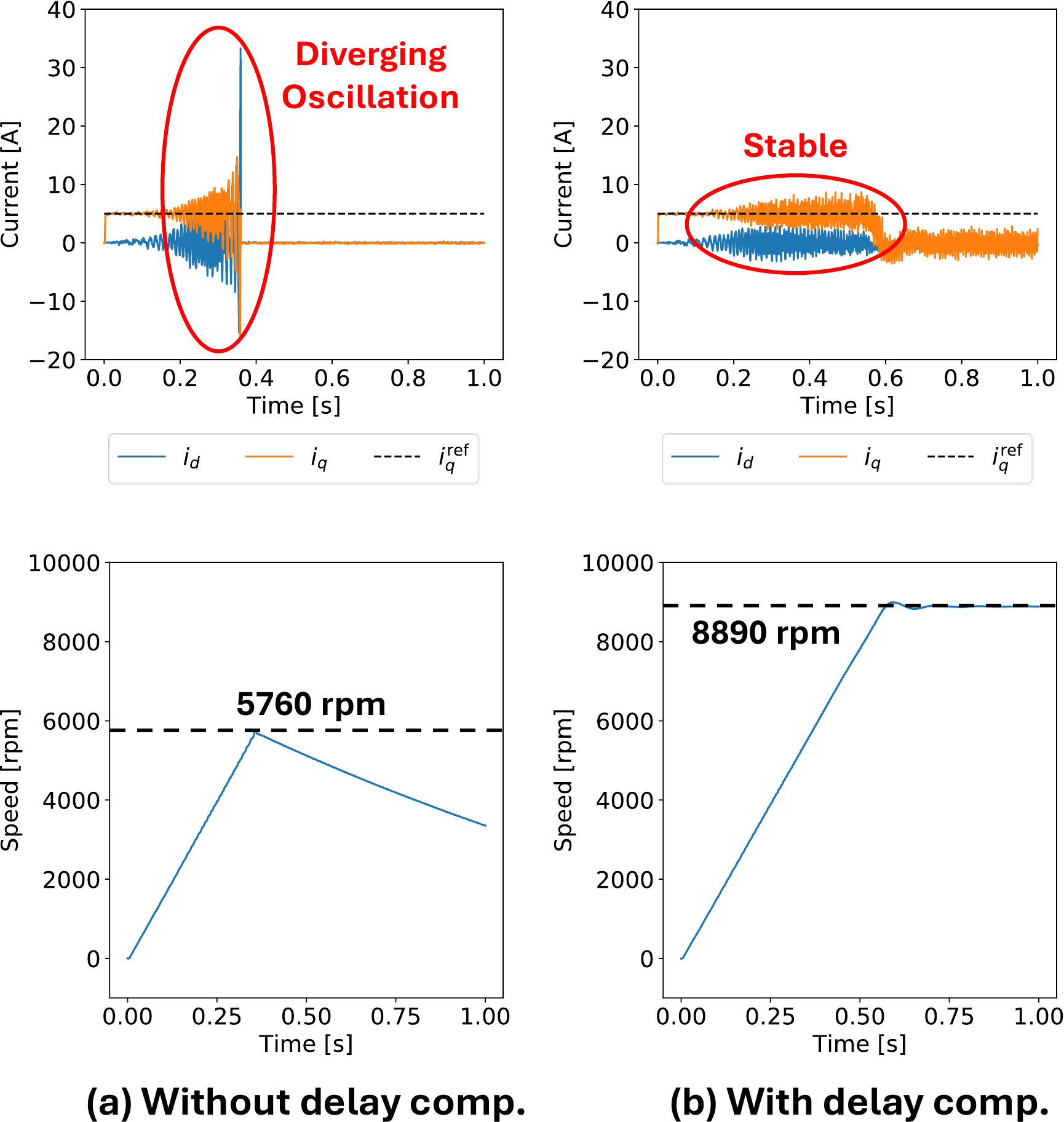}
  \caption{Current and speed responses at 96 V with and without delay compensation}
  \label{fig:speed_96V}
  \vspace{-1.0em}
\end{figure}

\section{Experiments} \label{sec:experiments}

\subsection{Basic Experiment: Switching Waveform Measurement} \label{subsec:basic_experiment}
In this experiment, switching waveforms were observed when \SI{10}{\ampere} was applied to the motor in order to verify proper switching operation. The oscilloscope used for the measurement was an MDO4054B-3 (Tektronix) with a bandwidth of \SI{500}{\mega\hertz}. The measurement results are shown in \figref{fig:waveform}.
Since the motor is a load with large current fluctuations, the slew rate was set to \SI{4.5}{\volt/\nano\second} as a conservative design choice. Although this value is slightly lower than typical GaN FET operation, it is still faster than conventional Si MOSFETs. The figure confirms that the ringing is sufficiently small.

\subsection{High-Speed Operation with Delay Compensation}
In this experiment, the improvement in motor speed was investigated when the supply voltage was increased from \SI{48}{\volt} to \SI{96}{\volt}. To evaluate the effect of delay compensation, comparative experiments were conducted with the delay compensation in the microcontroller enabled and disabled. A current command of \SI{5}{\ampere} was applied to the motor under no-load conditions, and the rotational speed was measured.

The results are summarized in \tabref{tab:speed}. Increasing the supply voltage from \SI{48}{\volt} to \SI{96}{\volt} increased the rotational speed from \SI{4340}{rpm} to \SI{8890}{rpm}. The speed at \SI{48}{\volt} remained below the specified no-load speed of \SI{5040}{rpm}, likely because of the \SI{95}{\percent} PWM duty-ratio limit and the modulation index. At \SI{48}{\volt}, both delay-compensation settings produced the same speed of \SI{4340}{rpm}. At \SI{96}{\volt}, enabling compensation increased the speed from \SI{5760}{rpm} to \SI{8890}{rpm}, a difference of \SI{3130}{rpm}.

The time-series waveforms of the current and speed at \SI{96}{\volt} are shown in \figref{fig:speed_96V}. Without delay compensation, interference between the d- and q-axes occurs due to control delay. The control becomes unstable and the system stops due to overcurrent detection. With delay compensation, stable operation is achieved. The electrical frequency reaches \SI{3110}{\hertz}, indicating extremely high-speed operation. Near the maximum speed, the current does not follow the command value due to speed saturation.

\begin{figure}[t]
  \centering
  \begin{minipage}[b]{0.45\columnwidth}
    \centering
    \includegraphics[width=\columnwidth]{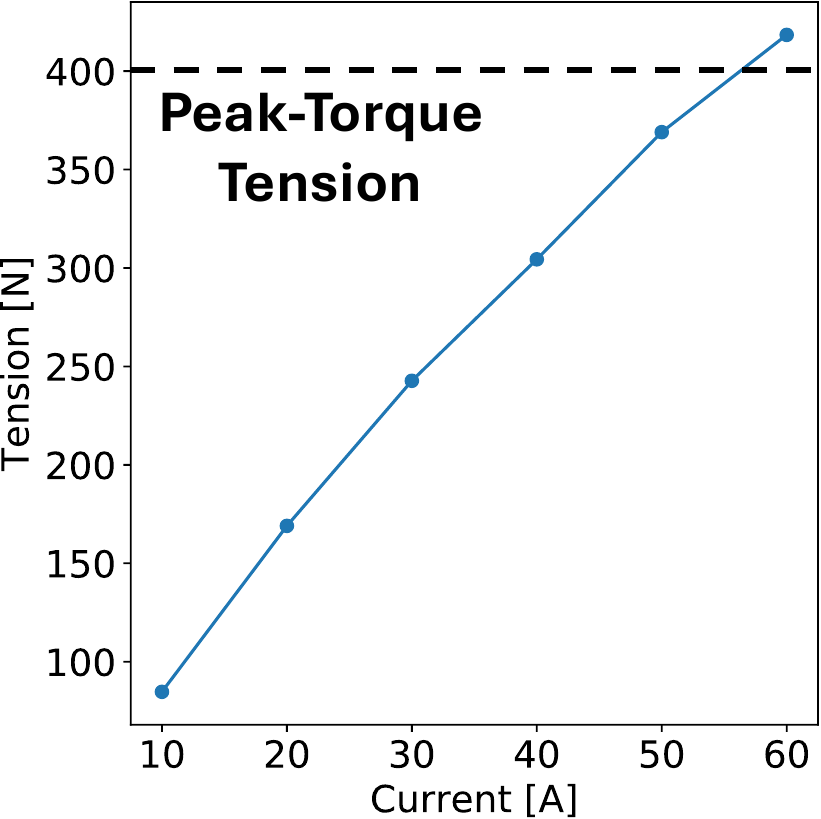}
    \vspace{-1.5em}
    \caption{Current--tension\\relationship}
    \label{fig:ro80_current_pulse}
  \end{minipage}
  \begin{minipage}[b]{0.45\columnwidth}
    \centering
    \includegraphics[width=\columnwidth]{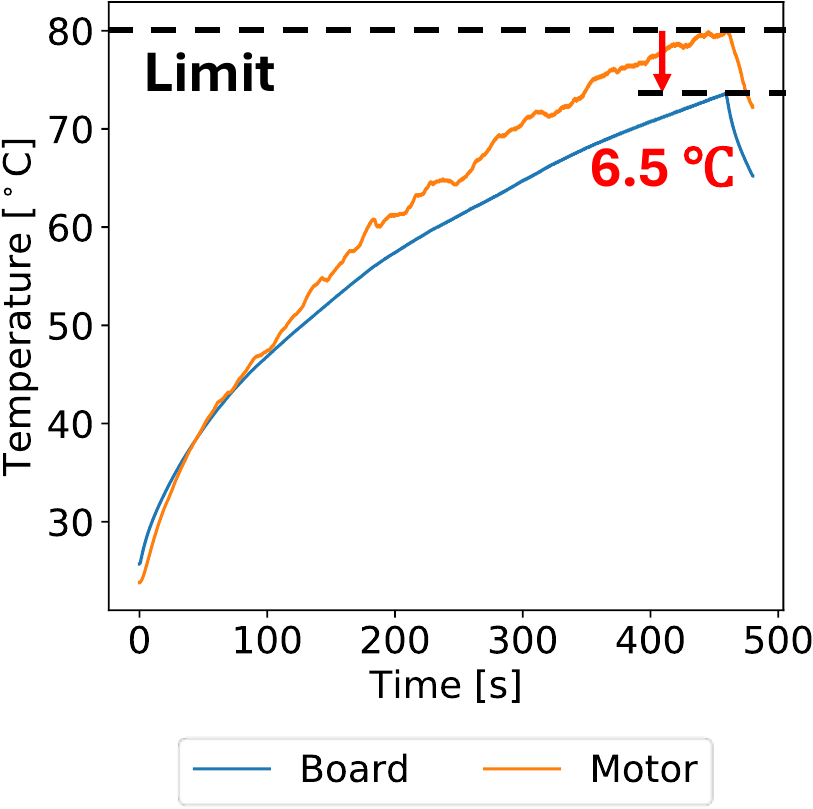}
    \vspace{-1.5em}
    \caption{Board and motor temperatures at 20 A}
    \label{fig:ro80_current_cont}
  \end{minipage}
  \vspace{-1.0em}
\end{figure}

\subsection{Validation of Sufficient Current Capability for the Target Motor}
In this experiment, it was verified that LunaDrive has sufficient peak and continuous current capability for the RO80 motor mounted on the fabricated wire module.

First, the peak current capability was evaluated. The driver capability was verified by confirming whether torque exceeding the peak torque specified in the motor datasheet could be generated. Since the definition of current, such as peak or RMS value, may differ depending on the specification, tension was directly measured in this experiment. The relationship between current and tension is shown in \figref{fig:ro80_current_pulse}. Tension was measured using a ZTS-500N (IMADA) with a maximum capacity of \SI{500}{N}. A current was applied for \SI{2}{s} under each condition, and up to \SI{60}{A} was supplied. As a result, the measured tension exceeded \SI{400}{N}, which is the value calculated from the specified peak torque. This confirms that LunaDrive has sufficient peak current capability.

Next, the continuous current capability was evaluated. A continuous current of \SI{20}{A} was applied to the motor, and the board and motor temperatures were compared to verify that the board was not the limiting factor. The experiment was conducted with the Flat Heat Sink attached. In this evaluation, FOC was performed with the electrical angle $\theta_e$ fixed at 0 so that the heat generation location remained constant.

When $\theta_e = 0$, the UVW phase currents with respect to the d-axis current $i_d$ are expressed as follows.

\begin{equation}
i_u = i_d,\quad
i_v = -\frac{1}{2} i_d,\quad
i_w = -\frac{1}{2} i_d
\end{equation}

Under this condition, the maximum current is continuously applied to the U phase. Therefore, heat generation is governed by the maximum current rather than the RMS value. In addition, the temperature sensor is located within 2~mm of the U-phase FET. This measurement condition is conservative with respect to the board temperature.

The results of the continuous current test are shown in \figref{fig:ro80_current_cont}. The temperature limit was set to 80~$^\circ$C. The results show that the motor reached the temperature limit first. At that time, the board temperature was 6.5~$^\circ$C lower than the motor temperature. After reaching the limit, the drive circuit was stopped for safety. These results confirm that LunaDrive has sufficient continuous current capability for the RO80 motor.

\begin{figure}[t]
  \centering
  \includegraphics[width=0.9\columnwidth]{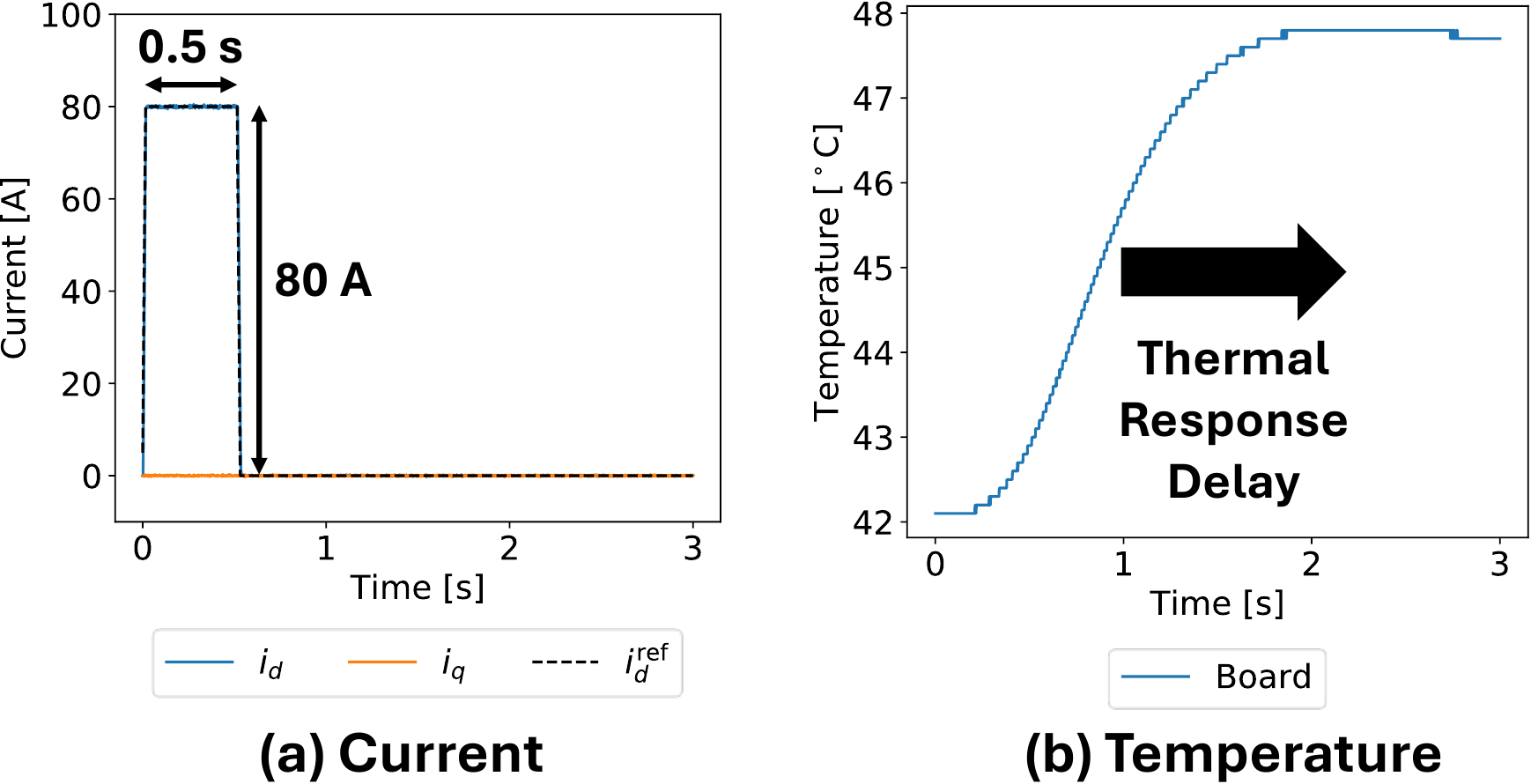}
  \caption{Current and board temperature when \SI{80}{\ampere} is commanded}
  \label{fig:U13_current_pulse}
\end{figure}

\begin{figure}[t]
  \centering
  \includegraphics[width=1.0\columnwidth]{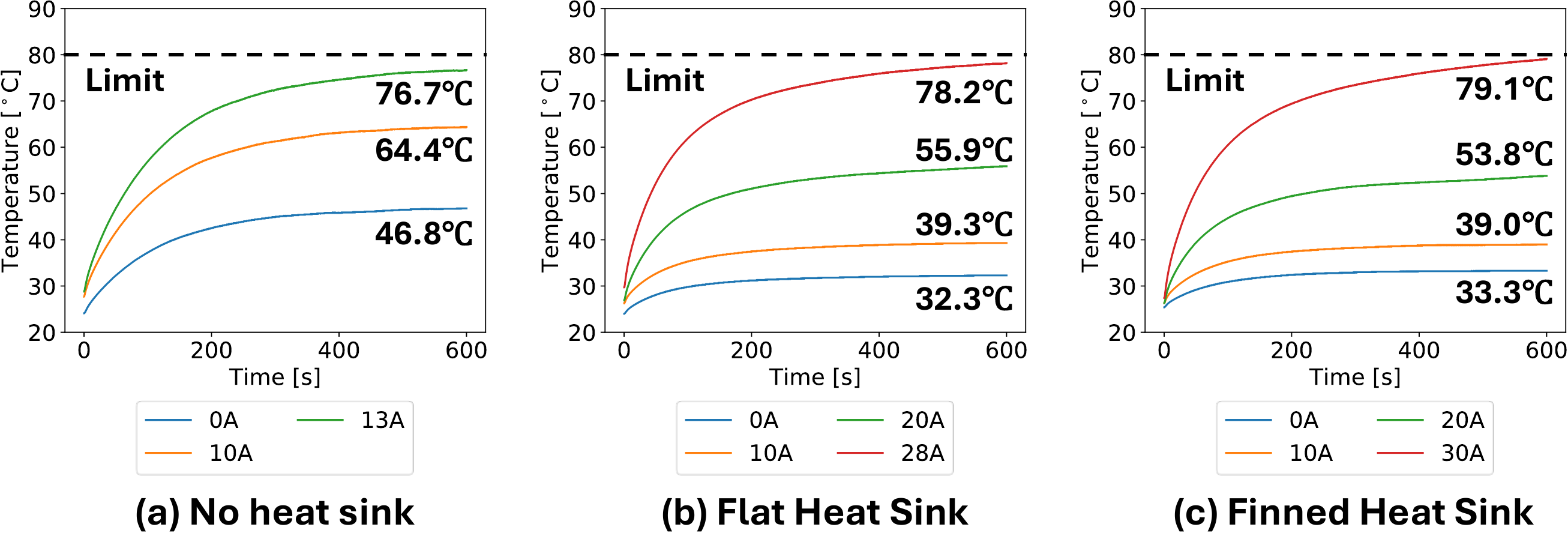}
  \vspace{-1.5em}
  \caption{Temperature rise over time at different continuous current levels}
  \label{fig:U13_current_cont}
  \vspace{-1.0em}
\end{figure}

\subsection{Characterization of the Peak and Continuous Current Capability}
In this experiment, the peak and continuous current capability of LunaDrive was investigated. Since the RO80 used in the wire module has a lower allowable current than LunaDrive, a U13II KV130 motor (T-Motor) for large drones was used as the load. As in the RO80 continuous current test, measurements were performed with the electrical angle $\theta_e$ fixed at 0. This is a conservative condition in which heat generation concentrates in a specific phase.

First, the peak current capability was evaluated. The experiment was conducted without a heat sink. The result when \SI{80}{A} was applied to the motor is shown in \figref{fig:U13_current_pulse}. It can be confirmed that \SI{80}{A} was applied for \SI{0.5}{s}. In addition, a large thermal response delay was observed in the board temperature. This suggests that, under large current conditions, the temperature of the GaN FET may not be measured accurately.

Next, the continuous current capability of LunaDrive was evaluated under different cooling conditions. Three conditions were tested: without a heat sink, with a Flat Heat Sink, and with a Finned Heat Sink.

The results are shown in \figref{fig:U13_current_cont}. Each current was applied for \SI{10}{min}. The current was increased in \SI{10}{A} increments at low temperatures and \SI{1}{A} increments near the \SI{80}{^\circ C} limit. The highest tested current below this limit was defined as the maximum continuous current. The resulting values were \SI{13}{A} without a heat sink, \SI{28}{A} with the Flat Heat Sink, and \SI{30}{A} with the Finned Heat Sink. Compared with the RO80 continuous current test, the lower temperature can be attributed to reduced heat transfer from the motor to the board due to lower motor heat generation.

In particular, without a heat sink, the board temperature increased to \SI{46.8}{^\circ C} even at \SI{0}{A}. This temperature rise is mainly attributed to switching losses caused by charging and discharging of the parasitic capacitances of the FETs and the snubber capacitors. Since these losses depend on the supply voltage, non-negligible heat generation occurs under the present condition of \SI{96}{V} operation.

At the maximum current condition, the temperature continued to rise even after \SI{10}{min}. However, \SI{80}{^\circ C} was set as a conservative limit considering safety margins. The board includes MLCCs with a temperature rating of \SI{125}{^\circ C}. Sufficient margin exists with respect to their rated temperature.

\subsection{High-Speed Load Motion Experiment}
\begin{figure}[t]
  \centering
  \includegraphics[width=1.0\columnwidth]{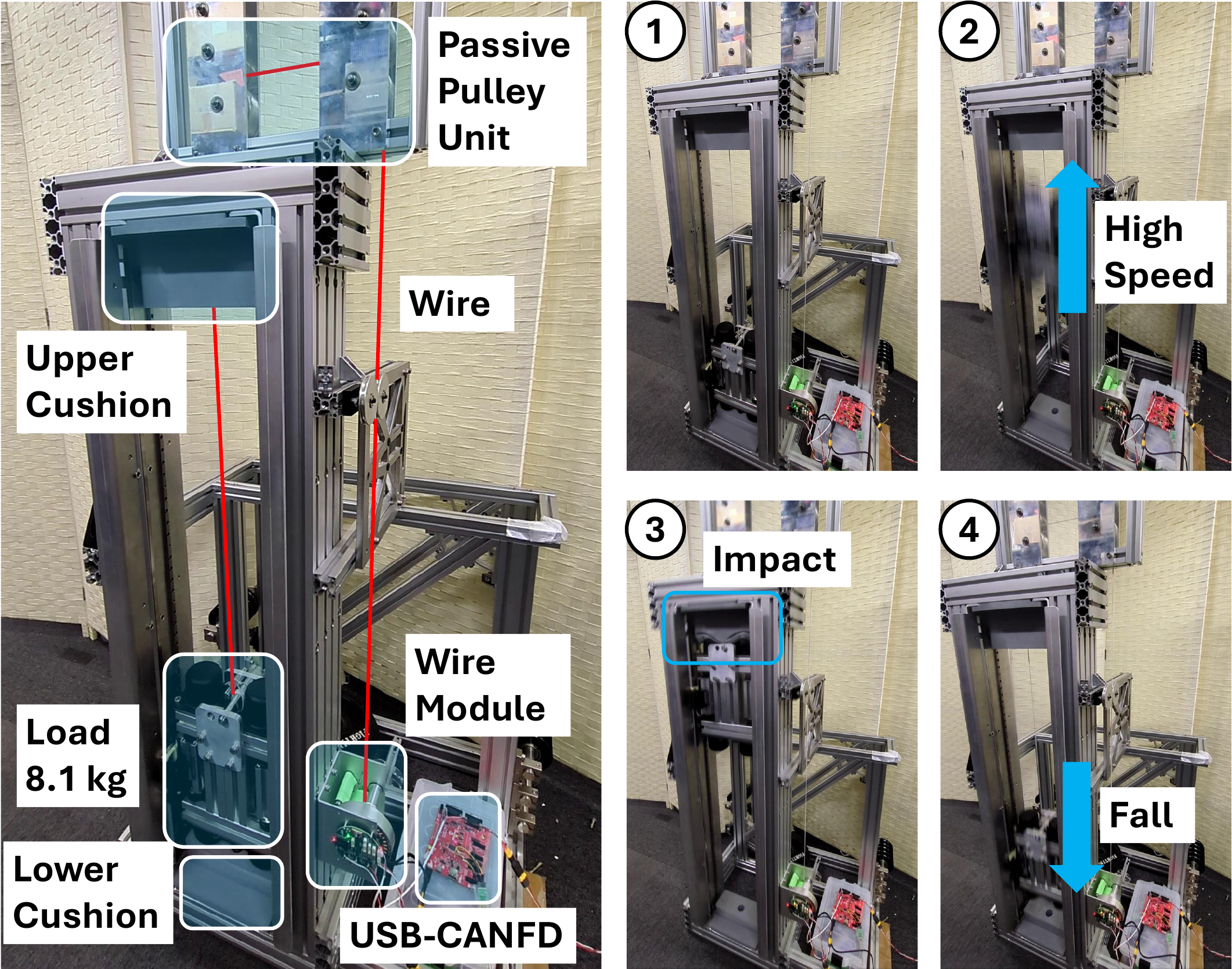}
  \vspace{-1.5em}
  \caption{Overview of the high-speed load lifting experiment. The sequential snapshots show the lifting motion at 100 V.}
  \label{fig:lift_motion}
\end{figure}

In this experiment, a high-speed load lifting test was conducted as an application experiment assuming a jumping robot. The setup is shown in \figref{fig:lift_motion}. The fabricated wire module was mounted on a universal wire testing machine \cite{temma2025tester}, and a load with a mass of \SI{8.1}{\kilogram} was lifted at high speed. The wire from the wire module was redirected through a passive pulley unit and then connected to the load. For safety, TPU cushions were placed above and below the moving part.

\begin{figure}[t]
  \centering
  \includegraphics[width=0.9\columnwidth]{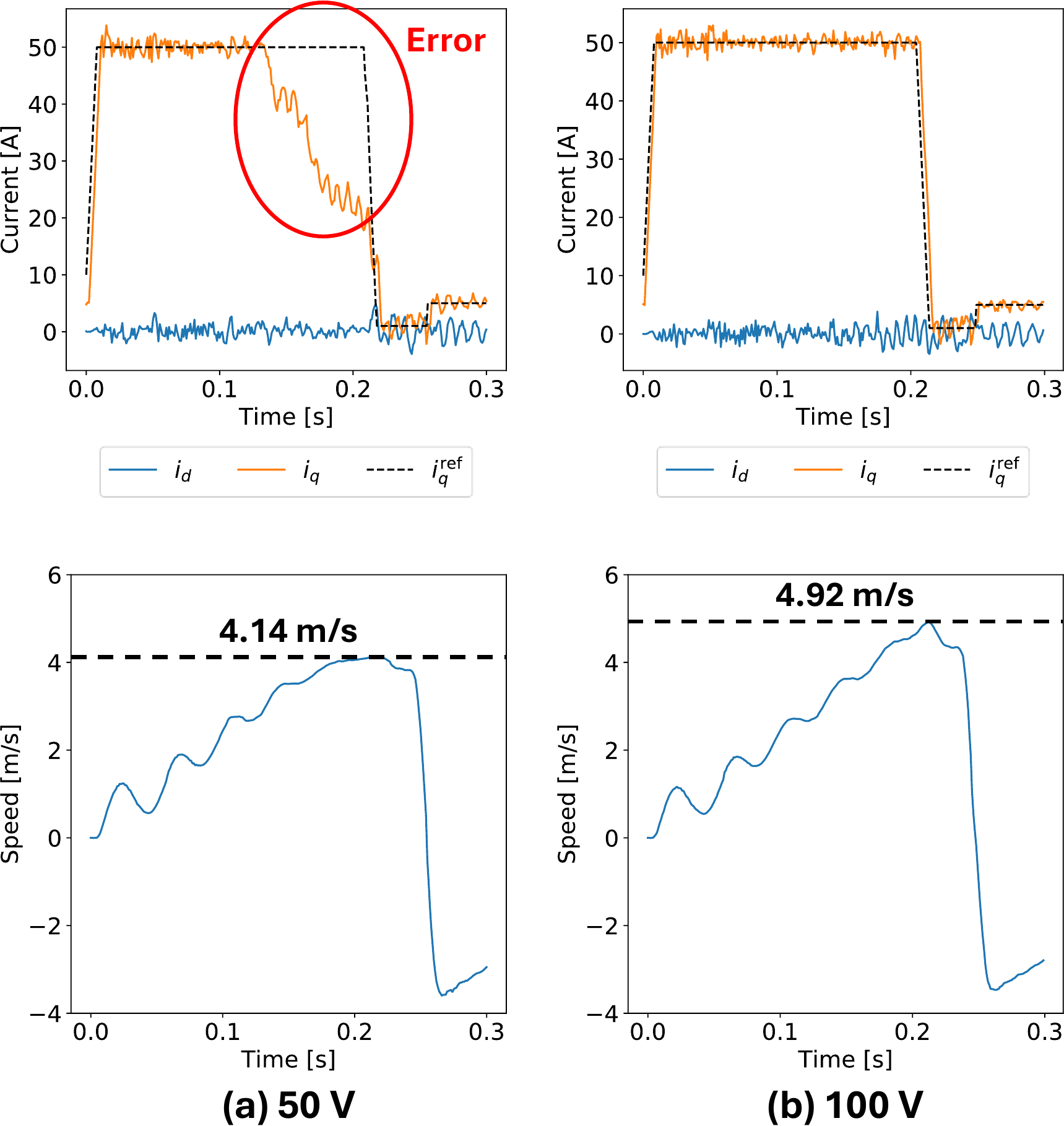}
  \caption{Current and speed responses during high-speed load lifting at 50 V and 100 V}
  \label{fig:lift}
  \vspace{-1.0em}
\end{figure}

Current control was implemented by commanding \SI{50}{\ampere} until \SI{0.5}{\meter} of wire was wound. Experiments were conducted under supply voltages of \SI{50}{\volt} and \SI{100}{\volt}. Since regenerative operation occurs during the falling motion of the load, a LiPo24S (\SI{100}{V}) battery was used instead of a power supply.

The results are shown in \figref{fig:lift}. Under the \SI{50}{\volt} condition, the current failed to follow the command in the latter half of acceleration due to speed saturation. In contrast, under the \SI{100}{\volt} condition, the current followed the command until just before the end of acceleration. The maximum speeds were \SI{4.14}{\meter/\second} and \SI{4.92}{\meter/\second}, corresponding to an improvement of approximately \SI{19}{\percent}. The limited improvement may be due to the short acceleration distance of \SI{0.5}{\meter}, which does not fully utilize the benefit of the higher voltage.

As shown in \figref{fig:lift_motion}, the load collided with the upper cushion at high speed and generated a large impact. This behavior indicates that LunaDrive enables high-current and high-voltage operation and can produce large instantaneous power.

\section{Discussion} \label{sec:discussion}
In this study, we proposed a motor driver, LunaDrive, which employs GaN FETs instead of Si MOSFETs to enable higher drive voltage and large-current operation for flat BLDC motors. In this section, the performance improvement obtained by adopting GaN FETs is discussed through comparison with an existing motor driver.

LunaDrive has a disk shape with a diameter of \SI{70}{mm}, excluding a part of the connector section, and a board thickness of \SI{8.51}{mm}. The rated voltage is \SI{96}{V}. The component rating and peak voltage are \SI{150}{V}. The continuous current was \SI{13}{A} without a heat sink at a thickness of \SI{8.51}{mm}, \SI{28}{A} with a Flat Heat Sink at a thickness of \SI{15}{mm}, and \SI{30}{A} with a Finned Heat Sink at a thickness of \SI{17.8}{mm}. A peak current of \SI{80}{A} was achieved.

Gold Solo Twitter (Elmo Motion Control, hereafter ELMO) is a compact high-power commercial motor driver. Its dimensions are 47.2×30×19.35 mm. The continuous current is \SI{3}{A} at 85 V operation without a heat sink at a thickness of \SI{19.35}{mm}, \SI{7}{A} with the FLAT Heat-Sink (ELMO) at a thickness of \SI{23.35}{mm}, and \SI{12}{A} with the FINs Heat-Sink (ELMO) at a thickness of \SI{28.85}{mm} \cite{gold_twitter_thermal_management}. The peak current is \SI{45}{A} for the R45/150 model, which can operate with LiPo24S at 100 V.

For flat motor applications, the thickness dimension is important. When compared based on thickness, LunaDrive shows advantages in both continuous current and peak current within the same voltage range. Since Gold Solo Twitter does not include an encoder, additional circuitry may be required in practical use, which may increase the thickness.

The heat sink of LunaDrive is fixed to the frame, and heat is dissipated through the frame. Therefore, a direct comparison of thermal conditions is not possible. However, LunaDrive without a heat sink shows higher continuous current capability than Gold Solo Twitter with a heat sink. LunaDrive also provides higher volumetric continuous current capability in the heat-sink-less configuration, which indicates a design advantage. For peak current, LunaDrive also achieves higher current capability when LiPo24S (\SI{100}{V}) operation is assumed.

These results show that the adoption of GaN FET enables both high-voltage and large-current operation.

\section{Conclusion} \label{sec:conclusion}
In this study, we proposed a compact motor driver, LunaDrive, which employs GaN FETs to drive high-power flat BLDC motors used in recent dynamic robots at high voltage beyond their nominal rating for high-speed operation. To address the large electrical angular velocity associated with high-voltage drive, delay compensation was implemented on both the encoder and microcontroller sides, and its effectiveness was verified.

LunaDrive has a compact structure with a diameter of \SI{70}{mm} and a thickness of \SI{8.51}{mm}. Large-current operation was achieved by reinforced wiring using a 2 oz 8-layer PCB and capacitor design based on effective capacitance. Evaluation of switching waveforms confirmed stable operation. A maximum speed of \SI{8890}{rpm}, corresponding to an electrical frequency of \SI{3110}{Hz}, was achieved. When the delay compensation was disabled, divergent oscillation occurred at approximately \SI{5760}{rpm}, which demonstrates the effectiveness of the compensation. The continuous current was \SI{13}{A} without a heat sink, \SI{28}{A} with a Flat Heat Sink, and \SI{30}{A} with a Finned Heat Sink. A peak current of \SI{80}{A} was achieved without a heat sink. Sufficient continuous and peak current capability was also confirmed for the RO80 installed in a wire module. Furthermore, experiments assuming actual robot operation demonstrated that velocity saturation can be avoided in the high-voltage and large-current region.

These results indicate that LunaDrive realizes high-voltage and large-current drive in a compact and thin configuration. This contributes to improving the motion performance of dynamic robots. Future work includes integration into an actual robot system to further validate its effectiveness.

{
  \bibliographystyle{IEEEtran}
  \bibliography{bib}
}

\end{document}